\documentclass[10pt,twocolumn,letterpaper]{article}

\usepackage{cvpr}              % To produce the CAMERA-READY version
\usepackage{graphicx}
\usepackage{amsmath}
\usepackage{amssymb}
\usepackage{booktabs}
\usepackage{multirow}
\usepackage[table,xcdraw]{xcolor}
\usepackage[pagebackref,breaklinks,colorlinks]{hyperref}
\usepackage{algorithm}
\usepackage{algpseudocode}

\usepackage[capitalize]{cleveref}
\crefname{section}{Sec.}{Secs.}
\Crefname{section}{Section}{Sections}
\Crefname{table}{Table}{Tables}
\crefname{table}{Tab.}{Tabs.}

\usepackage[normalem]{ulem}
\useunder{\uline}{\ul}{}
\def\cvprPaperID{7180} % *** Enter the CVPR Paper ID here
\def\confName{CVPR}
\def\confYear{2023}

\begin{document}

%%%%%%%%% TITLE - PLEASE UPDATE
\title{Block Selection Method for Using Feature Norm in Out-of-distribution Detection}

\author{Yeonguk Yu, Sungho Shin, Seongju Lee, Changhyun Jun, and Kyoobine Lee\\ 
Gwangju Institute of Science and Technology\\
{\tt\small \{yeon\_guk, hogili89, lsj2121, junch9634\}@gm.gist.ac.kr, kyoobinlee@gist.ac.kr}
% For a paper whose authors are all at the same institution,
% omit the following lines up until the closing ``}''.
% Additional authors and addresses can be added with ``\and'',
% just like the second author.
% To save space, use either the email address or home page, not both
%\and
%Second Author\\
%Institution2\\
%First line of institution2 address\\
%{\tt\small secondauthor@i2.org}
}
\maketitle

%   Detecting out-of-distribution (OOD) inputs during the inference stage is crucial for deploying the neural network in the real world. Previous methods commonly rely on the output of a network, which is known to be overconfident. In this paper, we argue that the block of the network can be deteriorated by overconfidence issue. To find a suitable block for OOD detection, we use jigsaw puzzle images generated from training images and see the ratio of feature map's norm between original image and jigsaw image. Our key idea is that the overconfident block will produce higher norm for jigsaw images compared to the one of suitable block. After then, we utilize a norm of the feature map from the suitable block as an indicator for OOD detection. We evaluate our method and demonstrate that our method outperform previous methods on CIFAR10 and ImageNet benchmarks. Finally, we show that by using the norm of the suitable block, the output of the model can be calibrated.

%%%%%%%%% ABSTRACT
\begin{abstract}
    Detecting out-of-distribution (OOD) inputs during the inference stage is crucial for deploying neural networks in the real world. Previous methods commonly relied on the output of a network derived from the highly activated feature map. In this study, we first revealed that a norm of the feature map obtained from the other block than the last block can be a better indicator of OOD detection. Motivated by this, we propose a simple framework consisting of \textbf{FeatureNorm}: a norm of the feature map and \textbf{NormRatio}: a ratio of FeatureNorm for ID and OOD to measure the OOD detection performance of each block. In particular, to select the block that provides the largest difference between FeatureNorm of ID and FeatureNorm of OOD, we create Jigsaw puzzle images as pseudo OOD from ID training samples and calculate NormRatio, and the block with the largest value is selected. After the suitable block is selected, OOD detection with the FeatureNorm outperforms other OOD detection methods by reducing FPR95 by up to 52.77\% on CIFAR10 benchmark and by up to 48.53\% on ImageNet benchmark. We demonstrate that our framework can generalize to various architectures and the importance of block selection, which can improve previous OOD detection methods as well.
\end{abstract}

%%%%%%%%% BODY TEXT
\section{Introduction}
\label{sec:intro}

% Background Knowledge about OOD detection 
Neural networks have widely been utilized in the real world, such as in autonomous cars\cite{pmlr-v119-filos20a, janai2020computer} and medical diagnoses\cite{pooch2020can, esteva2021deep}. In the real world, neural networks often encounter previously unseen input that are different from the training data. If the system fails to recognize those input as unknown input, there can be a dangerous consequence. For example, a medical diagnosis system may recognize an unseen 
disease image as one of the known diseases. This gives rise to the importance of the out-of-distribution (OOD) detection, which makes users operate a neural network system more safely in the real world.

In practice, various outputs of the network can be used as an indicator to separate the in-distribution (ID) and out-of-distribution (OOD) data. For instance, output probability\cite{hendrycks17baseline}, calibrated output probability\cite{liang2018enhancing}, and output energy\cite{liu2020energy} are used as an indicator. The output of a neural network is commonly calculated using a feature vector of the feature extractor and a weight vector of the classification layer. It is known that the norm of the feature vector can be an indicator of input image quality\cite{Parde2016DeepCN, ranjan2017l2, kim2022adaface} or level of awareness\cite{vaze2022openset}. Thus, we ask the following question: \textit{
Can we use the norm of the feature as an indicator to separate ID and OOD?}

% Observation and problem of the current network
% figure 1.
\begin{figure}[t]
    \centering
    \includegraphics[width=0.44\textwidth]{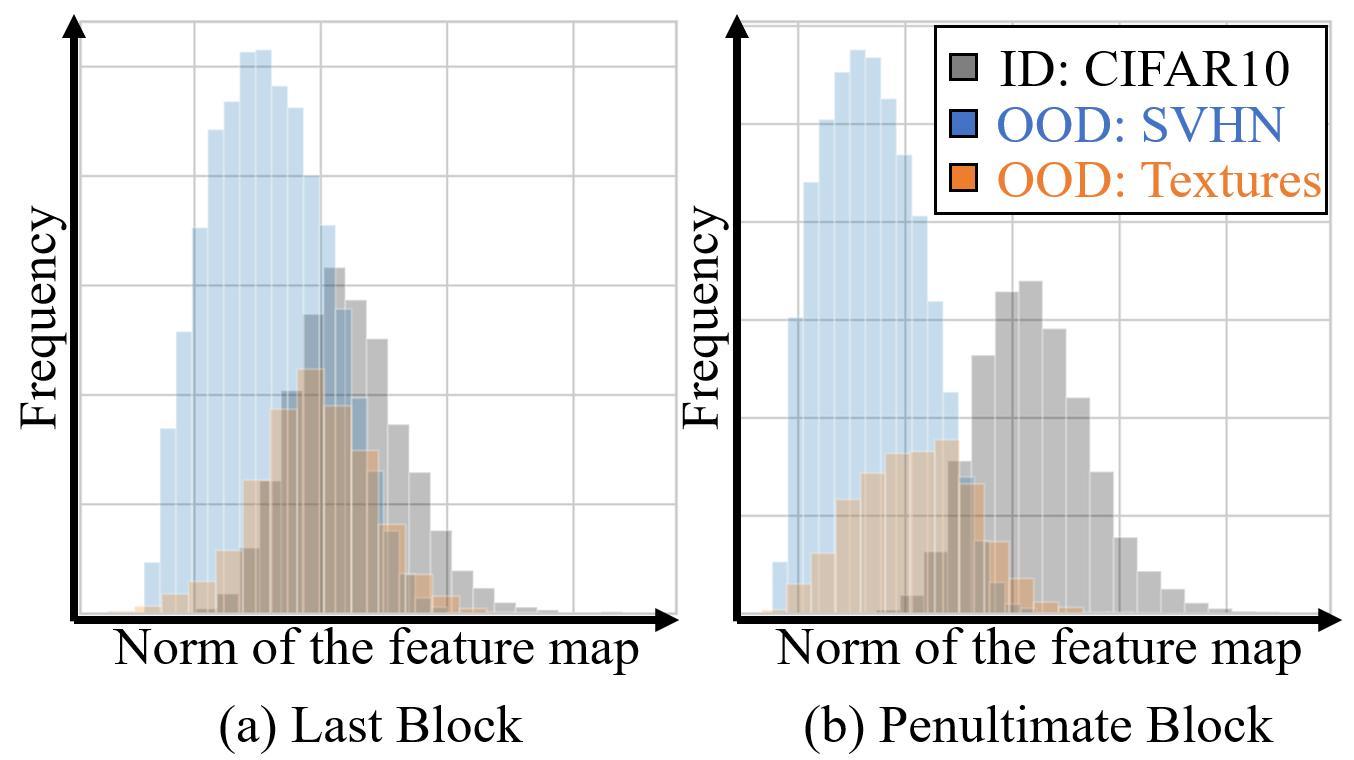}
    \caption{Histogram of norm of the feature map produced by convolutional blocks of ResNet18. In last block (a), the norm of ID (black) is hard to separate from OOD (blue, orange) compared to the one from the penultimate block (b).}
    \label{fig:observation}
\end{figure}

\begin{figure*}[t!]
    \centering
    \includegraphics[width=0.88\textwidth]{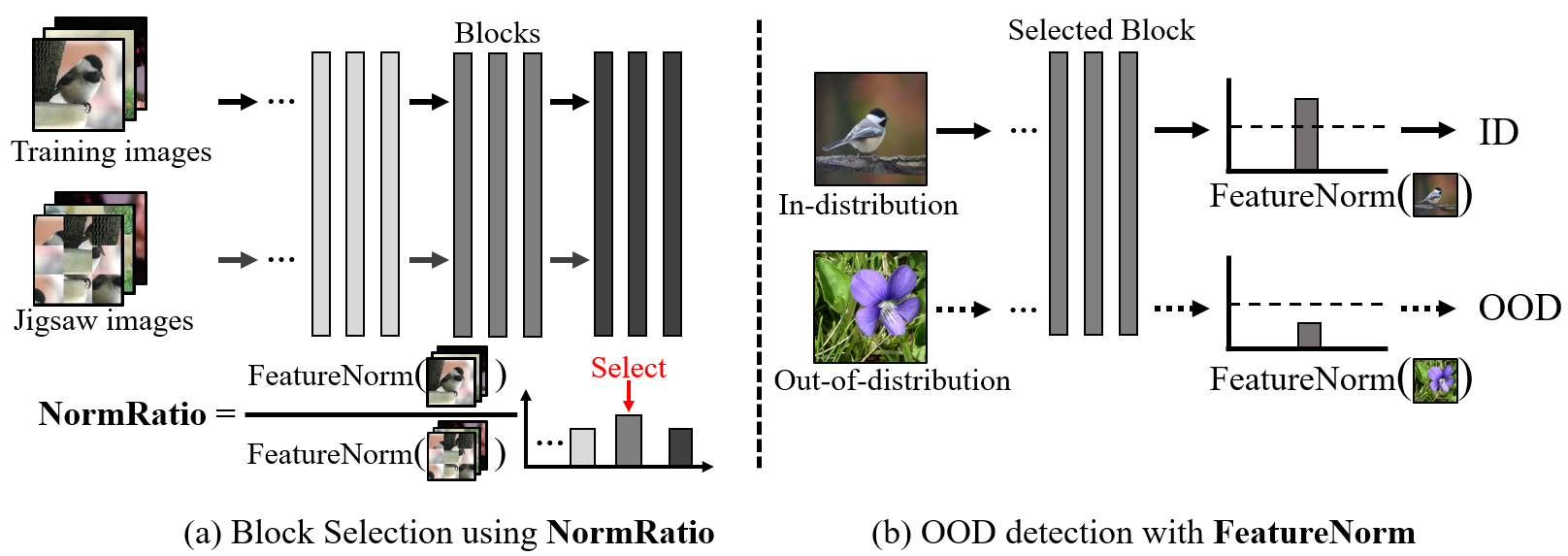}
    \caption{Illustration of our proposed out-of-distribution detection framework. FeatureNorm refers to a norm calculation for the given feature map produced by the block. We use \textit{NormRatio} of ID and pseudo OOD (i.e., Jigsaw puzzle images) to find which block is suitable for OOD detection (a). During inference time, for a given input image, the OOD score is calculated by \textit{FeatureNorm} on the selected block (b). If \textit{FeatureNorm} for a given input is smaller than the threshold, the given input is classified as OOD.}
    \label{fig:method}
\end{figure*}

In this paper, we first reveal the key observation concerning the last block of neural networks sometimes deteriorating owing to the overconfidence issue\cite{guo2017calibration, gupta2021calibration}. Empirically, we show that OOD images highly activate filters of the last block (i.e., large norm; see Figure \ref{fig:observation}, left) on a network trained with CIFAR10 while lowly activate filters of the penultimate block (i.e., small norm; see Figure \ref{fig:observation}, right). As a result, OOD detection methods that consider overactivated feature\cite{sun2021react} and overconfident output\cite{liang2018enhancing} have been successful. However, we find that the norm of the feature map for the OOD and ID is quite separable in the penultimate block compared to the last block.

% Brief description of our proposed method
This motivates a simple and effective OOD detection framework consists of (1) \textit{FeatureNorm}: the norm of the feature map and (2) \textit{NormRatio}: the ratio of \textit{FeatureNorm} for ID and OOD. In the case of a suitable block for OOD detection, the \textit{FeatureNorm} of ID is large, and the \textit{FeatureNorm} of OOD is small since its filters are trained to be activated on ID\cite{yosinski2015understanding, albawi2017understanding}. Thus, we use \textit{FeatureNorm} of the block that has the largest \textit{NormRatio} as an indicator to separate ID and OOD. Although \textit{NormRatio} directly represents the OOD detection performance of the block, we cannot access to the OOD before deployment. In order to select the block, that provides the largest difference between \textit{FeatureNorm} of ID and \textit{FeatureNorm} of OOD, we create pseudo OOD from ID training samples by generating Jigsaw puzzles\cite{noroozi2016unsupervised} and calculate \textit{NormRatio} with them, and the block with the largest value is selected without accessing OOD. Subsequently, we calculate \textit{FeatureNorm} of the given test image for OOD detection. The proposed OOD detection framework is shown in Figure \ref{fig:method}.

% Summary of the experiments and key results and contributions
We provide empirical and theoretical analysis of our proposed framework. We conduct experiments on common OOD detection benchmarks and show that our simple framework outperforms previous OOD detection methods. Below, we summarize our key results and contributions:
\begin{itemize}
    \item We introduce the \textit{FeatureNorm}, a norm of the feature map, and the \textit{NormRatio}, a ratio of FeatureNorm for ID and OOD to select a block for OOD detection. To the best of our knowledge, \textit{FeatureNorm} and \textit{NormRatio} are the first to explore and demonstrate the norm of the feature map can be an indicator of OOD detection.   
    
    \item We extensively evaluate our proposed framework on common benchmarks and establish state-of-the-art performances among post-hoc OOD detection methods. Our framework outperforms the best baseline by reducing FPR95 by up to 52.77\% on CIFAR10 benchmark and by up to 48.53\% on ImageNet benchmark.
    
    \item We provide ablation and theoretical analysis that improve understanding of our framework. Our analysis demonstrates an importance of norm from the suitable block, which can improve previous OOD detection methods. 
\end{itemize}
%------------------------------------------------------------------------

\section{Preliminaries}
We first describe the general setting of the supervised learning problem for the image classification network. In the general setting, the classification network is trained with cross-entropy loss for the given training dataset $D_{in} = \{(x_i, y_i)\}^I_{i=1}$, where $x_i \in \mathbb{R}^{3\times W\times H}$ is the input RGB image and $y_i \in \{1, 2, ..., K\}$ is the corresponding label with $K$ class categories. An OOD detection method is considered as post-hoc method if it does not modify anything during the training stage.

\paragraph{Out-of-distribution detection}
When deploying a network in the real world, a user can trust the network if the network classify known images correctly and detect the OOD image as "unknown". For an OOD detection problem with image classification networks, the given test image $x$ is considered as an OOD image when $x$ semantically (e.g., object identity) or non-semantically (e.g., camera settings or style of the image) differs from the images of the $D_{in}$. The decision of the OOD detection is a binary classification with a scoring function that produce ID-ness for the given image $x$. The main object of the OOD detection research is to find the scoring function that can effectively separate the ID samples and OOD samples.

\paragraph{Elements of convolutional neural networks}
Convolutional neural networks (CNNs) usually consist of a feature extractor and a classification layer. A feature extractor encodes an RGB image to a feature map with $M$ channel $z \in \mathbb{R}^{M\times W\times H}$ using its block, where $W$ and $H$ refer to the width and height of each feature map. Also, a classification layer encodes a feature map $z$ to a logit vector $v$. There have been various CNN architectures such as AlexNet\cite{krizhevsky2017imagenet}, VGG\cite{simonyan2014very}, ResNet\cite{he2016deep}, and MobileNet\cite{howard2019searching}. In this paper, we consider a block as a set of a single convolutional layer with an activation function in VGG architecture and a block as a residual block in ResNet and MobileNet.

Note that the output logit and the output probability of a CNN are commonly calculated as follows:
$$
v_i = W_i\cdot f = ||f||_2||W_i||_2\cos(\theta_i),
$$
$$
p_i = \frac{\exp(v_i)}{\sum_k \exp(v_k)},
$$
where $||\cdot||_2$, $v_i$, $f$, and $W_i$ denote the $L_2$-norm, $i$-th element of logit $v$, feature vector, and $i$-th class weight, respectively. Also, $\theta_i$ refers to the angle between the feature vector $f$ and $i$-th class weight vector $W_i$. Since the output probability distribution is calculated by applying a softmax function on the logit, larger $L_2$-norm feature and larger $L_2$-norm class weight produce a harder probability distribution\cite{xu2020feature}. Because the cross-entropy loss force the network to produce 1.0 probability for all training data, the norm of the feature vector and the norm of class weight are forced to be large.

%------------------------------------------------------------------------
\section{Method}
\label{sec:method}
\subsection{Overview of OOD detection framework}
Our OOD detection framework is based on the idea that the norm of feature map obtained from a suitable block for OOD detection can be a good indicator and the suitable block can be selected with ratio of ID and pseudo OOD which are generated from ID training samples. We illustrate our framework in Figure \ref{fig:method}. After the training is done, we select the block by \textit{NormRatio} for OOD detection (Figure \ref{fig:method}; left). Then, we use the norm of the feature map \textit{FeatureNorm} obtained from the selected feature map for OOD detection during the inference stage (Figure \ref{fig:method}; right). Specifically, we first generate the Jigsaw puzzle image as pseudo OOD from ID training samples and calculate \textit{NormRatio} of training samples and corresponding pseudo OOD. Since the Jigsaw puzzle images has destroyed object information, we argue that these images can be considered as OOD that is semantically shifted. Thus, \textit{NormRatio} of ID training samples and pseudo OOD (i.e., Jigsaw puzzle images) is suitable for finding the block that produce \textit{FeatureNorm} that can separate ID and OOD samples during the inference stage. Note that our proposed OOD detection framework does not modify the training stage of the network and once the input image is detected as in-distribution image during the inference stage, we can always obtain original output without having any disadvantage on classification accuracy.

\subsection{FeatureNorm: norm of the feature map}
We consider \textit{FeatureNorm}, a norm of the feature map, is an indicator of activation level of the block for the given image. In practice, we consider a pre-trained neural network for $K$-category image classification with a feature extractor and a classification layer. We denote by $z \in \mathbb{R}^{M \times W\times H}$ feature map obtained by the block of the feature extractor. The norm of each channel of feature map $z_i \in \mathbb{R}^{1 \times W\times H}$ is calculated as follows:
\begin{equation}
    \label{eq:featurenorm}
    a_i = \sqrt{\sum_w^W \sum_h^H \text{max}(z_i(w, h), 0)^2},
\end{equation}
where $z_i(w, h)$ is the $w$-th, $h$-th element of the feature map $z_i$. This equation can be interpreted as Frobenius norm of the rectified $z_i$ by ReLU function. We utilize a ReLU function to eliminate the effect of negative elements of feature map, which can be seemed as deactivation of filters. Thus, the $a_i$ represents the level of activation of $i$-th channel for the obtained feature map $z$.

Subsequently, the channel-wise averaged norm of the feature map for the block $B$ is calculated as follows:
\begin{equation}
    \label{eq:featurenorm2}
    f^\text{\textit{FeatureNorm}}(x; B) = \frac{1}{M}\sum_{m=1}^M a_m,
\end{equation}
where $f^\text{\textit{FeatureNorm}}(x; B) \in \mathbb{R}$ is the level of activation for the given image $x$ and the block $B$. During the inference stage, OOD detection using \textit{FeatureNorm} with a suitable block $B_s$ can be conducted with a threshold $\gamma$: 
\begin{equation}
G(x; \theta)=
\begin{cases}
\text{ID} & $if $ f^\text{\textit{FeatureNorm}}(x; B_s) \geq \gamma \\
\text{OOD} & $else$,
\end{cases}
\end{equation}
where the threshold $\gamma$ is typically chosen so that 95\% of ID data is correctly classified as ID (i.e., true positive rate of 95\%) and the $\theta$ refers to the neural network. 

\subsection{NormRatio: measure of block's suitability}
\label{sec:normratio}
We consider the \textit{NormRatio}, a ratio of ID \textit{FeatureNorm} and pseudo OOD \textit{FeatureNorm}, is an indicator of block's suitability for OOD detection. \textit{NormRatio} directly represents suitability of the block for OOD detection since the suitable block will produce large ID \textit{FeatureNorm} and the small OOD \textit{FeatureNorm}. In practice, the main problem of the \textit{NormRatio} for selecting a block is that we cannot access the OOD before deployment. Thus, we need to generate the pseudo OOD, that can represent the OOD may seem during the inference stage, to calculate \textit{NormRatio}. We argue that using the \textit{NormRatio} for selecting the block with pseudo OOD that can represent the most hard OOD can achieve the best OOD detection results, and since the semantically shifted OOD images are known to be the hardest OOD to detect\cite{hsu2020generalized}, we generate the 3$\times$3 Jigsaw puzzle image, which is semantically shifted, as done in \cite{noroozi2016unsupervised} using training samples. Our selection algorithm using \textit{NormRatio} is described as Algorithm \ref{alg:selection}.

\begin{algorithm}[h]
\caption{Block selection using \textit{NormRatio}}
\label{alg:selection}
\label{alg:selection}
\begin{algorithmic}
\State\textbf{Input:} Block list $\{B_1, ... B_N\}$, training data $\{x_i, y_i\}^I_{i=1}$
\While{$B_n \in \{B_1, ... B_N\}$}
    \While{$i \in I$}
\State $\blacktriangleright$\textbf{Create jigsaw image:}
        \State {$\hat{x_i} \leftarrow Jigsaw(x_i)$}
\State $\blacktriangleright$\textbf{Compute NormRatio:}
    \State $R_{(n,i)} = \frac{f^{FeatureNorm}(x_i; B_n)}{f^{FeatureNorm}(\hat{x}_i; B_n)}$
    \EndWhile
    \State $R_n \leftarrow \frac{1}{I} \Sigma_i R_{(n,i)}$
\EndWhile
    \State $s \leftarrow \text{argmax}_n(R_n)$
    \State {\textbf{Return} $B_s$}

\end{algorithmic}
\end{algorithm}

%------------------------------------------------------------------------
\section{Experiments}
\label{sec:experiemnts}

% In this section, we evaluate OOD detection performance of proposed \textit{FeatureNorm} with the block selection using \textit{NormRatio} of ID and pseudo OOD. Pseudo OOD are 3$\times$3 Jigsaw puzzle images that generated from each ID training sample. After selecting the block for each model with \textit{NormRatio}, we evaluate on a small-scale OOD benchmark based on CIFAR10\cite{krizhevsky2009learning}. Subsequently, we evaluate on a large-scale OOD benchmark that is recently introduced\cite{huang2021mos} based on ImageNet\cite{imagenet_cvpr09}. 

\paragraph{Setup} We use commonly utilized CNN architectures: ResNet18\cite{he2016deep}, VGG11\cite{simonyan2014very} and WideResNet\cite{BMVC2016_87} with depth 28 and width 10 (WRN28) for the CIFAR10\cite{krizhevsky2009learning} benchmark. The ResNet18 and VGG11 are trained with batch size 128 for 100 epochs with weight decay of 0.0005. The WRN28 is trained with batch size 128 for 200 epochs with weight decay of 0.0005. In all training, the SGD optimizer with momentum 0.9 and initial learning rate of 0.1 is utilized, except VGG11 use initial learning rate of 0.05. The learning rate is decreased by a factor of 10 at 50, 75, 90 training epochs for ResNet18 and VGG11, and at 100, 150 training epochs for WRN28. Also, we use pretrained ResNet50\cite{he2016deep}, VGG16\cite{simonyan2014very}, and MobileNetV3\_large\cite{howard2019searching} (MobileNetV3) architectures provided by Pytorch for ImageNet\cite{imagenet_cvpr09} benchmark. At test time, all images are resize to 32$\times$32 for CIFAR10 networks, and to 256$\times$256 and center crop to size of 224$\times$224 for ImageNet networks. We use SVHN\cite{netzer2011reading}, Textures\cite{cimpoi14describing}, LSUN-crop\cite{yu15lsun}(LSUN(c)), LSUN-resize\cite{yu15lsun}(LSUN(r)), iSUN\cite{isun}, and Places365\cite{7968387} as OOD datasets for CIFAR10 benchmark. We use iNaturalist\cite{inatural}, SUN\cite{5539970}, PLACES\cite{7968387}, and Textures\cite{cimpoi14describing}, which are sampled by Huang \etal\cite{huang2021mos}, as OOD dataset For the ImageNet benchmark.

\paragraph{Evaluation Metrics} We measure the quality of OOD detection using the two most widely adopted metrics in OOD detection researches: which are (1) area under the receiver operating characteristic curve (AUROC; \cite{davis2006relationship, fawcett2006introduction}) and (2) false positive rate at 95\% true positive rate (FPR95; \cite{liang2018enhancing}). AUROC plots the true positive rate of ID data against false positive rate of OOD data by varying a OOD detection threshold. Thus, it can represent the probability that ID samples will have a higher score than OOD samples. FPR95 is a false positive rate at threshold is set to produce 95\% true positive rate. Therefore, it can represent OOD detection performance when an application requirement is recall of 95\%. In summary, the higher AUROC and the lower FPR95 represent the better quality of the OOD detection method.

\paragraph{Comparison with previous methods} We compare our framework with other post-hoc OOD detection method which calculate OOD score from a model trained on ID data using cross-entropy loss. Although, ODIN\cite{liang2018enhancing} requires validation set of ID and OOD for hyperparameter setting, we set these hyperparameters without OOD data as in \cite{hsu2020generalized} for fair comparison without access to OOD. As a result, we compare our method with Maximum Softmax Probability (MSP; \cite{hendrycks17baseline}), ODIN\cite{liang2018enhancing}, Energy\cite{liu2020energy}, Energy+ReAct\cite{sun2021react}, and Energy+DICE\cite{sun2022dice}.

\begin{table}[t!]
\centering
\resizebox{0.48\textwidth}{!}{
\huge
\begin{tabular}{@{}llccc@{}}
\toprule
\multicolumn{1}{c}{\textbf{ID}} & \textbf{Architecture} & \textbf{Selected Block Name} & \textbf{Output Size} & \textbf{Depth} \\ \midrule
\multirow{3}{*}{CIFAR10} & ResNet18 & Block 4.1 & 512 $\times$ 4 $\times$ 4 & N-1 \\
 & WRN28 & Block 3.3 & 640 $\times$ 8 $\times$ 8 & N-1 \\
 & VGG11 & Layer 7 & 512 $\times$ 4 $\times$ 4 & N-2 \\ \hline
\multirow{3}{*}{ImageNet} & ResNet50 & Block 4.2 & 2048 $\times$ 7 $\times$ 7 & N-1 \\
 & VGG16 & Layer 13 & 512 $\times$ 14 $\times$ 14 & N \\
 & MobileNetV3 & Block 17 & 960 $\times$ 7 $\times$ 7 & N \\ \bottomrule
\end{tabular}}
\caption{Summary of the selected blocks for each architecture. Depth N represents the last block, while Depth 1 represents the first block.}
\label{tab:BlockSelection}
\end{table}

\paragraph{Block selection for OOD detection using \textit{NormRatio}} we evaluate OOD detection performance of proposed \textit{FeatureNorm} with the block selection using \textit{NormRatio} of ID and pseudo OOD. Pseudo OOD are 3$\times$3 Jigsaw puzzle images that generated from each ID training sample. Once the training of the network is done, we calculate the \textit{NormRatio} for every block using Algorithm \ref{alg:selection}. Since the various architectures are used for the experiments, we summarize results of block selection in Table \ref{tab:BlockSelection}. We find that our method choose the block for each architecture consistently. For instance, the Block 4.1, Block 3.3 and Layer 7 are selected for ResNet18, WRN28 and VGG11 in all five trials.

%------------------------------------------------------------------------
\begin{table*}[hbt!]
\centering
\resizebox{0.95\textwidth}{!}{%
\huge
\begin{tabular}{@{}clcccccccccccccc@{}}
\toprule
 & \multicolumn{1}{c}{} & \multicolumn{12}{c}{\textbf{OOD}} & \multicolumn{2}{c}{} \\ \cline{3-14}
 & \multicolumn{1}{c}{} & \multicolumn{2}{c}{SVHN} & \multicolumn{2}{c}{Textures} & \multicolumn{2}{c}{LSUN©} & \multicolumn{2}{c}{LSUN®} & \multicolumn{2}{c}{iSUN} & \multicolumn{2}{c}{Places365} & \multicolumn{2}{c}{\multirow{-2}{*}{\textbf{Average}}} \\ \cline{3-16}
\multirow{-3}{*}{\textbf{Architecture}} & \multicolumn{1}{c}{\multirow{-3}{*}{\textbf{Method}}} & FPR95↓ & AUROC↑ & FPR95↓ & AUROC↑ & FPR95↓ & AUROC↑ & FPR95↓ & AUROC↑ & FPR95↓ & AUROC↑ & FPR95↓ & AUROC↑ & FPR95↓ & AUROC↑ \\ \midrule
 & MSP\cite{hendrycks17baseline} & 52.12 & 92.20 & 59.47 & 89.56 & 32.83 & 95.62 & 48.35 & 93.07 & 50.30 & 92.58 & 60.70 & 88.42 & 50.63 & 91.91 \\
 & ODIN\cite{liang2018enhancing} & 33.83 & 93.03 & {\ul 45.49} & 90.01 & 7.29 & 98.62 & \textbf{20.05} & \textbf{96.56} & \textbf{23.09} & \textbf{96.01} & 45.06 & {\ul 89.86} & {\ul 29.14} & 94.02 \\
 & Energy\cite{liu2020energy} & 30.47 & 94.05 & 45.83 & {\ul 90.37} & 7.21 & 98.63 & {\ul 23.62} & {\ul 95.93} & 27.14 & 95.34 & \textbf{43.67} & \textbf{90.29} & 29.66 & {\ul 94.10} \\
 & Energy+ReAct\cite{sun2021react} & 40.54 & 90.54 & 48.61 & 88.44 & 15.12 & 96.86 & 27.01 & 94.74 & 30.57 & 93.95 & {\ul 44.99} & 89.37 & 34.47 & 92.32 \\
 & Energy+DICE\cite{sun2022dice} & {\ul 25.95} & {\ul 94.66} & 47.22 & 89.82 & {\ul 3.83} & {\ul 99.26} & 27.70 & 95.01 & 31.07 & 94.42 & 49.28 & 88.08 & 30.84 & 93.54 \\
\multirow{-6}{*}{\textbf{ResNet18}} & \cellcolor[HTML]{EFEFEF}FeatureNorm (ours) & \cellcolor[HTML]{EFEFEF}\textbf{7.13} & \cellcolor[HTML]{EFEFEF}\textbf{98.65} & \cellcolor[HTML]{EFEFEF}\textbf{31.18} & \cellcolor[HTML]{EFEFEF}\textbf{92.31} & \cellcolor[HTML]{EFEFEF}\textbf{0.07} & \cellcolor[HTML]{EFEFEF}\textbf{99.96} & \cellcolor[HTML]{EFEFEF}27.08 & \cellcolor[HTML]{EFEFEF}95.25 & \cellcolor[HTML]{EFEFEF}{\ul 26.02} & \cellcolor[HTML]{EFEFEF}{\ul 95.38} & \cellcolor[HTML]{EFEFEF}62.54 & \cellcolor[HTML]{EFEFEF}84.62 & \cellcolor[HTML]{EFEFEF}\textbf{25.67} & \cellcolor[HTML]{EFEFEF}\textbf{94.36} \\ \hline
 & MSP\cite{hendrycks17baseline} & 42.10 & {\ul 91.85} & 53.30 & {\ul 87.45} & 24.85 & 96.37 & 37.81 & 93.71 & 40.11 & 93.05 & 50.73 & 88.58 & 41.49 & 91.84 \\
 & ODIN\cite{liang2018enhancing} & 37.08 & 88.36 & 47.58 & 82.85 & 6.14 & 98.65 & {\ul 20.51} & {\ul 95.04} & {\ul 22.95} & {\ul 94.22} & {\ul 41.03} & 86.57 & 29.22 & 90.95 \\
 & Energy\cite{liu2020energy} & {\ul 33.11} & 90.54 & {\ul 46.06} & 85.09 & 5.86 & 98.76 & 22.68 & 94.90 & 25.12 & 94.17 & \textbf{39.08} & {\ul 88.50} & {\ul 28.65} & {\ul 91.99} \\
 & Energy+ReAct\cite{sun2021react} & 98.31 & 39.94 & 91.85 & 60.80 & 96.76 & 57.11 & 77.63 & 80.15 & 79.48 & 78.67 & 73.29 & 77.98 & 86.22 & 65.78 \\
 & Energy+DICE\cite{sun2022dice} & 37.84 & 86.99 & 50.77 & 79.70 & {\ul 2.54} & {\ul 99.43} & 26.30 & 92.89 & 28.30 & 92.14 & 43.46 & 84.65 & 31.53 & 89.30 \\
\multirow{-6}{*}{\textbf{WRN28}} & \cellcolor[HTML]{EFEFEF}FeatureNorm (ours) & \cellcolor[HTML]{EFEFEF}\textbf{3.83} & \cellcolor[HTML]{EFEFEF}\textbf{99.18} & \cellcolor[HTML]{EFEFEF}\textbf{14.23} & \cellcolor[HTML]{EFEFEF}\textbf{97.06} & \cellcolor[HTML]{EFEFEF}\textbf{0.32} & \cellcolor[HTML]{EFEFEF}\textbf{99.81} & \cellcolor[HTML]{EFEFEF}\textbf{8.13} & \cellcolor[HTML]{EFEFEF}\textbf{98.32} & \cellcolor[HTML]{EFEFEF}\textbf{5.98} & \cellcolor[HTML]{EFEFEF}\textbf{98.71} & \cellcolor[HTML]{EFEFEF}48.69 & \cellcolor[HTML]{EFEFEF}\textbf{90.91} & \cellcolor[HTML]{EFEFEF}\textbf{13.53} & \cellcolor[HTML]{EFEFEF}\textbf{97.33} \\ \hline
 & MSP\cite{hendrycks17baseline} & 68.07 & 90.02 & 63.86 & 89.37 & 46.63 & 93.73 & 70.19 & 86.29 & 71.81 & 85.71 & 68.08 & 87.25 & 64.77 & 88.73 \\
 & ODIN\cite{liang2018enhancing} & 53.84 & 92.23 & 48.09 & 91.94 & 19.95 & 97.01 & 54.29 & 89.47 & 56.61 & 88.87 & 52.34 & 89.86 & 47.52 & {\ul 91.56} \\
 & Energy\cite{liu2020energy} & 53.13 & 92.26 & {\ul 47.04} & {\ul 92.08} & 18.51 & {\ul 97.20} & {\ul 53.02} & {\ul 89.58} & {\ul 55.39} & {\ul 88.97} & {\ul 51.67} & {\ul 89.95} & {\ul 46.46} & \textbf{91.67} \\
 & Energy+ReAct\cite{sun2021react} & 58.81 & 83.28 & 51.73 & 87.47 & 23.40 & 94.77 & \textbf{47.19} & \textbf{89.68} & \textbf{51.30} & \textbf{88.07} & \textbf{50.47} & \textbf{87.39} & 47.15 & 88.44 \\
 & Energy+DICE\cite{sun2022dice} & {\ul 47.81} & {\ul 93.27} & 50.95 & 91.77 & {\ul 16.73} & 97.06 & 64.26 & 87.83 & 65.83 & 87.43 & 59.23 & 88.53 & 50.80 & 90.98 \\
\multirow{-6}{*}{\textbf{VGG11}} & \cellcolor[HTML]{EFEFEF}FeatureNorm (ours) & \cellcolor[HTML]{EFEFEF}\textbf{8.84} & \cellcolor[HTML]{EFEFEF}\textbf{98.24} & \cellcolor[HTML]{EFEFEF}\textbf{24.62} & \cellcolor[HTML]{EFEFEF}\textbf{95.11} & \cellcolor[HTML]{EFEFEF}\textbf{3.38} & \cellcolor[HTML]{EFEFEF}\textbf{99.36} & \cellcolor[HTML]{EFEFEF}71.17 & \cellcolor[HTML]{EFEFEF}83.12 & \cellcolor[HTML]{EFEFEF}62.80 & \cellcolor[HTML]{EFEFEF}86.05 & \cellcolor[HTML]{EFEFEF}65.25 & \cellcolor[HTML]{EFEFEF}85.20 & \cellcolor[HTML]{EFEFEF}\textbf{39.34} & \cellcolor[HTML]{EFEFEF}91.18 \\ \bottomrule
\end{tabular}}
\caption{Performance of OOD detection on CIFAR10 benchmarks. All methods in the table has no access to OOD data during training and validation. The best and second-best results are indicated in \textbf{bold} and \underline{underline}, respectively. All values are percentages averaged over five runs.}
\label{tab:cifar10}
\end{table*}

\section{Results}
\label{sec:results}
\subsection{Result on CIFAR10 benchmark}
In Table \ref{tab:cifar10}, we report the performance of OOD detection for ResNet18, WRN28, and VGG11 architectures using various post-hoc detection methods. The performance are calculated using FPR95 and AUROC on six OOD datasets. Our proposed method achieved the best average performance on both ResNet18 and WRN28, and best FPR95 on VGG11. Note that our method reduces the average FPR95 by 13.45\%, 52.77\%, and 15.33\% compared to the second best results on ResNet18, WRN28, and VGG11, respectively. 

As shown in Table \ref{tab:cifar10}, our method consistently outperforms other method on three OOD datasets: SVHN, Textures and LSUN(c). Also, we find that our method is weaker to LSUN(r), iSUN, and Places365. We argue that our method is stronger to detect images from SVHN, Textures and LSUN(c) since its image has low complexity compare to CIFAR10\cite{lin2021mood} and activation of the image cumulatively differs from the early stage of the network to later stage (see \ref{sec:effect_norm}). In contrast, LSUN(r) and iSUN have large complexity\cite{lin2021mood}, which makes its activation large on the shallow layer and to be detected more easily when using a deeper architecture (i.e., WRN28). Finally, Places365 has similar complexity as CIFAR10\cite{lin2021mood} which can be interpreted as images from Places365 have similar low-level abstraction information and semantically shifted compared to ID (i.e., semantically shifted OOD\cite{hsu2020generalized}).

\subsection{Result on ImageNet benchmark}
\begin{table*}[hbt!]
\centering
\resizebox{0.95\textwidth}{!}{%
\begin{tabular}{@{}clcccccccccc@{}}
\toprule
 & \multicolumn{1}{c}{} & \multicolumn{8}{c}{\textbf{OOD}} & \multicolumn{2}{c}{} \\ \cline{3-10}
 & \multicolumn{1}{c}{} & \multicolumn{2}{c}{iNaturalist} & \multicolumn{2}{c}{SUN} & \multicolumn{2}{c}{PLACES} & \multicolumn{2}{c}{Textures} & \multicolumn{2}{c}{\multirow{-2}{*}{\textbf{Average}}} \\ \cline{3-12} 
\multirow{-3}{*}{\textbf{Architecture}} & \multicolumn{1}{c}{\multirow{-3}{*}{\textbf{Method}}} & \multicolumn{1}{l}{FPR95↓} & \multicolumn{1}{l}{AUROC↑} & \multicolumn{1}{l}{FPR95↓} & \multicolumn{1}{l}{AUROC↑} & \multicolumn{1}{l}{FPR95↓} & \multicolumn{1}{l}{AUROC↑} & \multicolumn{1}{l}{FPR95↓} & \multicolumn{1}{l}{AUROC↑} & \multicolumn{1}{l}{FPR95↓} & \multicolumn{1}{l}{AUROC↑} \\ \midrule
 & MSP†\cite{hendrycks17baseline} & 54.99 & 87.74 & 70.83 & 80.86 & 73.99 & 79.76 & 68.00 & 79.61 & 66.95 & 81.99 \\
 & ODIN†\cite{liang2018enhancing} & 47.66 & 89.66 & 60.15 & 84.59 & 67.89 & 81.78 & 50.23 & 85.62 & 56.48 & 85.41 \\
 & Energy†\cite{liu2020energy} & 55.72 & 89.95 & 59.26 & 85.89 & 64.92 & 82.86 & 53.72 & 85.99 & 58.41 & 86.17 \\
 & Energy+ReAct†\cite{sun2021react} & \textbf{20.38} & \textbf{96.22} & \textbf{24.20} & \textbf{94.20} & \textbf{33.85} & \textbf{91.58} & 47.30 & 89.80 & \textbf{31.43} & \textbf{92.95} \\
 & Energy+DICE†\cite{sun2022dice} & 25.63 & 94.49 & {\ul 35.15} & {\ul 90.83} & {\ul 46.49} & {\ul 87.48} & {\ul 31.72} & {\ul 90.30} & {\ul 34.75} & 90.78 \\
\multirow{-6}{*}{\textbf{ResNet50}} & \cellcolor[HTML]{EFEFEF}FeatureNorm (Ours) & \cellcolor[HTML]{EFEFEF}{\ul 22.01} & \cellcolor[HTML]{EFEFEF}{\ul 95.76} & \cellcolor[HTML]{EFEFEF}42.93 & \cellcolor[HTML]{EFEFEF}90.21 & \cellcolor[HTML]{EFEFEF}56.80 & \cellcolor[HTML]{EFEFEF}84.99 & \cellcolor[HTML]{EFEFEF}\textbf{20.07} & \cellcolor[HTML]{EFEFEF}\textbf{95.39} & \cellcolor[HTML]{EFEFEF}35.45 & \cellcolor[HTML]{EFEFEF}{\ul 91.59} \\ \hline
 & MSP\cite{hendrycks17baseline} & 56.72 & 87.26 & 75.66 & 78.31 & 77.89 & 77.60 & 64.84 & 81.66 & 68.78 & 81.21 \\
 & ODIN\cite{liang2018enhancing} & {\ul 42.66} & {\ul 92.13} & 61.31 & 86.51 & 67.33 & 83.87 & 44.57 & 89.82 & 53.97 & 88.08 \\
 & Energy\cite{liu2020energy} & 44.60 & 91.77 & 59.34 & {\ul 86.82} & {\ul 66.27} & {\ul 83.95} & 43.90 & 89.94 & {\ul 53.53} & {\ul 88.12} \\
 & Energy+ReAct\cite{sun2021react} & 99.94 & 34.50 & 99.87 & 35.01 & 99.25 & 37.54 & 96.45 & 49.12 & 98.88 & 39.04 \\
 & Energy+DICE\cite{sun2022dice} & 49.70 & 90.03 & {\ul 58.42} & 86.71 & 68.97 & 83.04 & {\ul 38.95} & {\ul 90.66} & 54.01 & 87.61 \\
\multirow{-6}{*}{\textbf{VGG16}} & \cellcolor[HTML]{EFEFEF}FeatureNorm (Ours) & \cellcolor[HTML]{EFEFEF}\textbf{16.78} & \cellcolor[HTML]{EFEFEF}\textbf{96.69} & \cellcolor[HTML]{EFEFEF}\textbf{28.09} & \cellcolor[HTML]{EFEFEF}\textbf{94.37} & \cellcolor[HTML]{EFEFEF}\textbf{41.78} & \cellcolor[HTML]{EFEFEF}\textbf{90.21} & \cellcolor[HTML]{EFEFEF}\textbf{23.53} & \cellcolor[HTML]{EFEFEF}\textbf{95.05} & \cellcolor[HTML]{EFEFEF}\textbf{27.55} & \cellcolor[HTML]{EFEFEF}\textbf{94.08} \\ \hline
 & MSP\cite{hendrycks17baseline} & 56.04 & 87.31 & 74.19 & 79.08 & 77.03 & 78.23 & 65.00 & 81.64 & 68.07 & 81.57 \\
 & ODIN\cite{liang2018enhancing} & {\ul 39.93} & {\ul 93.10} & {\ul 55.22} & {\ul 87.87} & 64.11 & {\ul 85.09} & {\ul 38.28} & {\ul 91.24} & {\ul 49.39} & {\ul 89.33} \\
 & Energy\cite{liu2020energy} & 54.04 & 91.15 & 68.13 & 85.89 & 69.37 & 83.91 & 54.91 & 88.88 & 61.61 & 87.46 \\
 & Energy+ReAct\cite{sun2021react} & 40.98 & 91.17 & 59.82 & 84.80 & {\ul 63.07} & 81.53 & 58.78 & 85.17 & 55.66 & 85.67 \\
 & Energy+DICE\cite{sun2022dice} & 60.94 & 84.72 & 63.4 & 82.7 & 75.88 & 77.88 & 42.98 & 87.36 & 60.80 & 83.17 \\
\multirow{-6}{*}{\textbf{MobileNetV3}} & \cellcolor[HTML]{EFEFEF}FeatureNorm (Ours) & \cellcolor[HTML]{EFEFEF}\textbf{33.10} & \cellcolor[HTML]{EFEFEF}\textbf{92.71} & \cellcolor[HTML]{EFEFEF}\textbf{42.41} & \cellcolor[HTML]{EFEFEF}\textbf{88.60} & \cellcolor[HTML]{EFEFEF}\textbf{58.46} & \cellcolor[HTML]{EFEFEF}\textbf{81.79} & \cellcolor[HTML]{EFEFEF}\textbf{8.60} & \cellcolor[HTML]{EFEFEF}\textbf{98.26} & \cellcolor[HTML]{EFEFEF}\textbf{35.64} & \cellcolor[HTML]{EFEFEF}\textbf{90.34} \\ \bottomrule
\end{tabular}}
\caption{Performance of OOD detection on ImageNet benchmarks. All methods in the table has no access to OOD data during training and validation.  The best and second-best results are indicated in \textbf{bold} and \underline{underline}, respectively. All values are obtained over a single run with the pretrained network provided by Pytorch. † indicates the result is reported by Sun \etal\cite{sun2022dice}.}
\label{tab:imagenet}
\end{table*}

In Table \ref{tab:imagenet}, we report the performance of OOD detection for ResNet50, VGG16, and MobileNetV3 architectures. The performance are calculated using FPR95 and AUROC on four datasets. Our proposed method achieved the best averaged performance on VGG16 and MobileNetV3 architectures. Note that our method reduces the FPR95 on ImageNet benchmark by 48.53\% and 27.84\% compared to the second best results when using VGG16 and MobileNetV3 architectures. In contrast, we find that our method does not effective on ResNet50 architecture compared to the other methods. We argue that the block structure with batch normalization layer of ResNet reduced the separation gap between ID and OOD samples (see \ref{sec:overconfidence}).

Note that our method consistently outperforms other methods on detecting Textures dataset. We argue that images of Textures are far-OOD\cite{yang2022openood} and have a lot of low-level complexity images compared to the other OOD dataset, and the activation cumulatively differs from the early stage of the network to later stage (see \ref{sec:effect_norm}). Also, iNaturalist usually have higher complexity images compared to ImageNet. As a result, deep networks can detect the iNaturalist as OOD unlike VGG11 and ResNet18 on CIFAR10 benchmark. Finally, SUN and PLACES have similar complexity level compared to ImageNet, which means that OOD images is semantically shifted and hard to be detected\cite{hsu2020generalized}. 

%------------------------------------------------------------------------
\section{Discussion}
\label{sec:discussion}

\subsection{Effect of NormRatio}
\label{sec:effect_ratio}
\begin{figure}[t]
    \centering
    \includegraphics[width=0.48\textwidth]{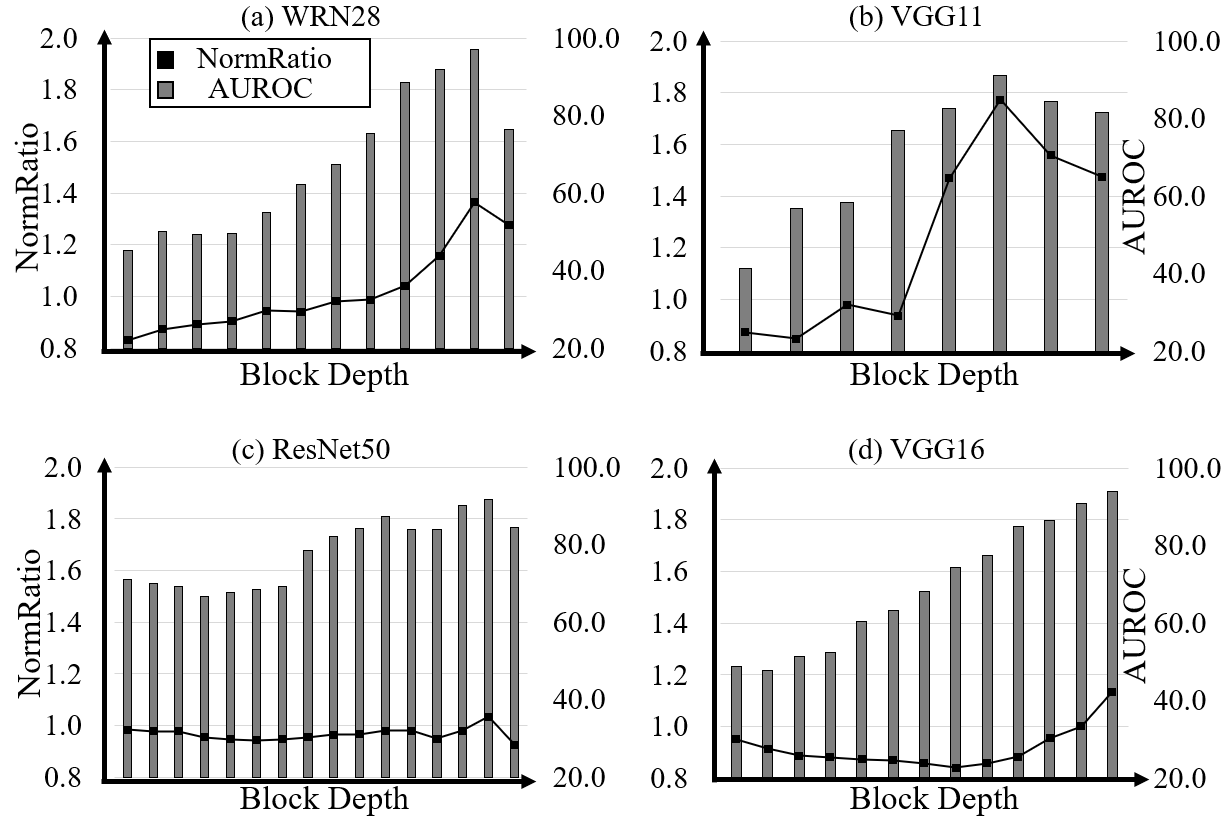}
    \caption{\textit{NormRatio} using ID and pseudo OOD calculated on each block (solid line with square marker) and the performance of OOD detection with \textit{FeatureNorm} on each block (gray bar). It shows that the best OOD detection performance is achieved when using the block that produce the largest \textit{NormRatio}.}
    \label{fig:normratio}
\end{figure}

We calculate the \textit{NormRatio} using the ID training samples and the pseudo OOD, which is a jigsaw puzzle generated from the ID image, to select the suitable block for OOD detection. Our insight is that the block that has the largest \textit{NormRatio} is the suitable for detecting pseudo OOD (i.e., jigsaw puzzle) and the other OOD can be seen during the inference stage as well. To find out that the \textit{NormRatio} can represent the OOD detection performance of the block, we calculate the \textit{NormRatio} and OOD detection performance for each block. In Figure \ref{fig:normratio}, we show that the best OOD detection performance can be achieved by the block that produce the largest \textit{NormRatio}. Also, in Figure \ref{fig:normblocks}, we show that \textit{FeatureNorm} for the given ID (black), pseudo OOD (gray) and various OOD (SVHN: red, Places: blue, LSUN(r): orange). We see that the \textit{FeatureNorm} of pseudo OOD (gray) acts as the \textit{FeatureNorm} of OOD images that has enough low-level abstraction (blue, orange).

\subsection{Effect of FeatureNorm}
\label{sec:effect_norm}
\begin{figure}[t]
    \centering
    \includegraphics[width=0.42\textwidth]{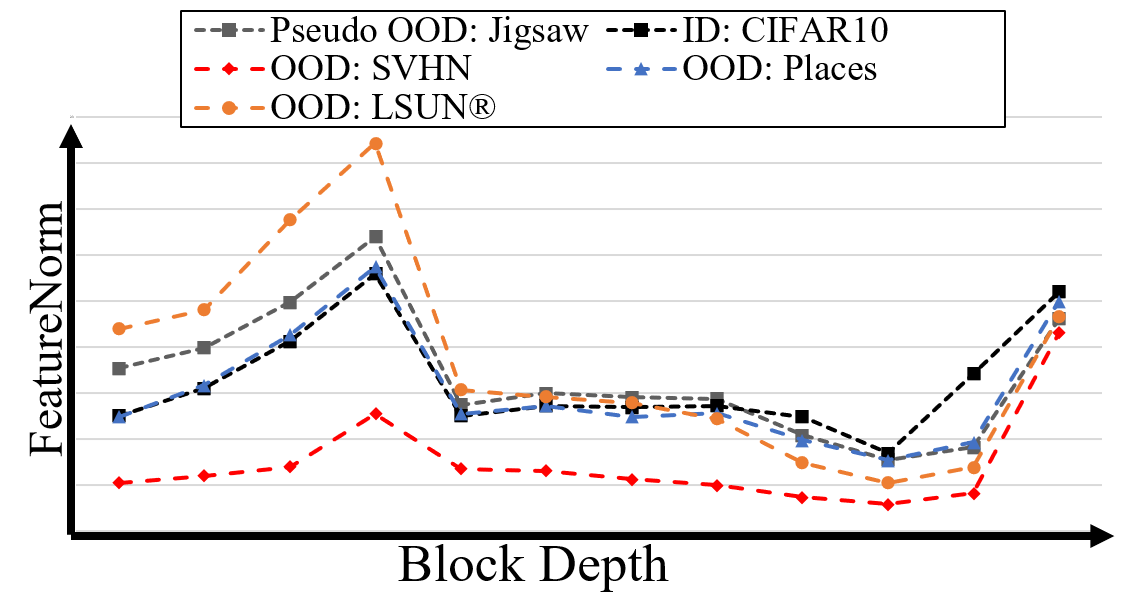}
    \caption{The \textit{FeatureNorm} calculated using every block for the given ID (black), Pseudo OOD (gray), and OOD images (SVHN: red, Places: blue, LSUN(r): orange). We show that the high complexity image (LSUN(r)) activates shallow block and low complexity image (SVHN) does not activate the block except the last one. We find that the jigsaw puzzle image represents the high complexity OOD image with no semantic information.}
    \label{fig:normblocks}
\end{figure}

\begin{figure}[t]
    \centering
    \includegraphics[width=0.45\textwidth]{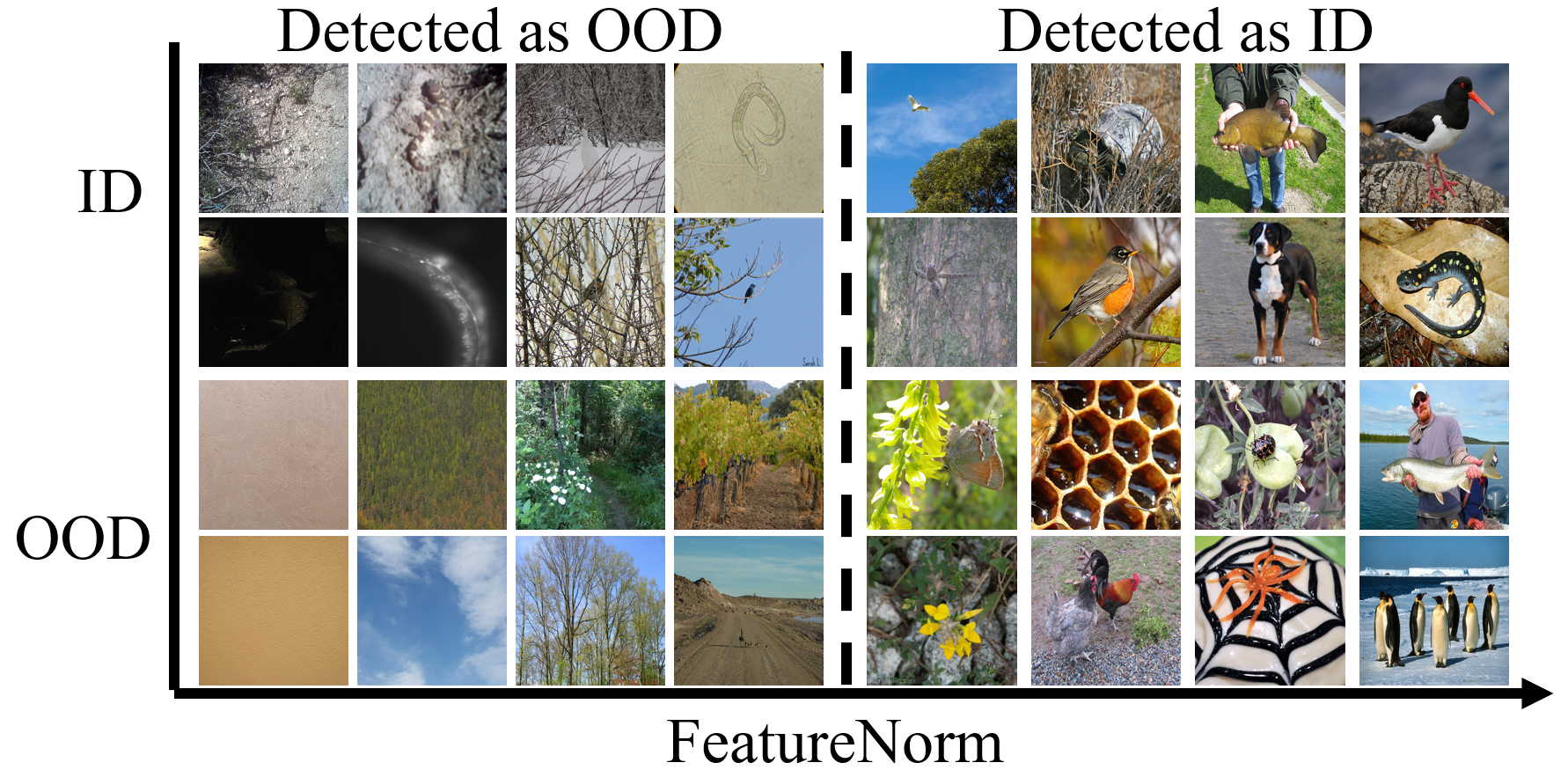}
    \caption{The example results of our proposed OOD detection framework with a VGG16 architecture for ImageNet. We demonstrate that the low-complexity images obtain low \textit{FeatureNorm}, while high-complexity images obtain high \textit{FeatureNorm}. Note that, some OOD images actually contain ID objects (e.g., tench, king penguin, honeycomb).}
    \label{fig:results}
\end{figure}

To demonstrate the effectiveness of \textit{FeatureNorm} as an indicator for OOD detection, we show the change of \textit{FeatureNorm} of various input through blocks in Figure \ref{fig:normblocks}. In Figure \ref{fig:normblocks}, we demonstrate that norm of the OOD images with low-complexity (SVHN) is low consistently on all blocks except the last one. On the other way, \textit{FeatureNorm} of the OOD images with high complexity (LSUN(r)) is higher than ID in shallow blocks since the shallow block of network act as edge detector\cite{albawi2017understanding, yosinski2015understanding} and the high level complexity image tends to have large low-level abstraction. Also, high complexity images obtain large \textit{FeatureNorm} in shallow blocks, the \textit{FeatureNorm} is reduced since it cannot activate deeper blocks that act as high-level abstraction detector. We argue that network like VGG11 or ResNet18 is hard to separate ID and OOD with high-level complexity since it has a few deep blocks compared to WRN28. As a result, the low quality ID image obtains low \textit{FeatureNorm} and the high quality ID image or OOD image with high-level semantic information obtains high \textit{FeatureNorm} as shown in Figure \ref{fig:results}.

\subsection{Output calibration using the selected block}
The output of the network is deteriorated when the last block of the network produces a high norm for both ID and OOD, and we argue that the network output can be calibrated by replacing the norm of the last block with the norm of the selected block. We evaluate the OOD detection method on ResNet18, WRN28, and VGG11, which are suffered by overconfident last block, with and without replacing. In the Table \ref{tab:normcorrection}, we show that the existing OOD detection methods that utilize the output of the network can be improved by selecting a suitable norm for the OOD detection. This suggests that the current network is underperforming in OOD detection by the overconfident last block, and can be calibrated by the block selection. 

\begin{table}[]
\centering
\resizebox{0.45\textwidth}{!}{%
\begin{tabular}{@{}clcccc@{}}
\toprule
\multirow{2}{*}{\textbf{Architecture}} & \multicolumn{1}{c}{\multirow{2}{*}{\textbf{Method}}} & \multicolumn{2}{c}{\textbf{w/o the selected norm}} & \multicolumn{2}{c}{\textbf{w/ the selected norm}} \\ \cmidrule(l){3-6} 
 & \multicolumn{1}{c}{} & FPR95↓ & AUROC↑ & FPR95↓ & AUROC↑ \\ \midrule
\multirow{3}{*}{ResNet18} & MSP & 50.63 & 91.91 & 35.56 & 94.66 \\
 & ODIN & 29.14 & 94.02 & 15.99 & 97.03 \\
 & Energy & 29.66 & 94.10 & 17.59 & 96.69 \\ \hline
\multirow{3}{*}{WRN28} & MSP & 41.49 & 91.84 & 38.38 & 93.59 \\
 & ODIN & 29.22 & 90.95 & 23.71 & 94.75 \\
 & Energy & 28.65 & 91.99 & 24.01 & 94.87 \\ \hline
\multirow{3}{*}{VGG11} & MSP & 64.77 & 88.73 & 56.57 & 90.55 \\
 & ODIN & 47.52 & 91.56 & 35.62 & 93.50 \\
 & Energy & 46.46 & 91.67 & 35.42 & 93.58 \\ \bottomrule
\end{tabular}}
\caption{An ablation study with and without selected norm for other previous OOD detection methods. We demonstrate that replacing the norm of the last block with the norm of the selected block improve the performance of previous methods. The results are averaged over five runs.}
\label{tab:normcorrection}
\end{table}

\subsection{Structure of the block cause the overconfidence}
\label{sec:overconfidence}
Despite the effect of \textit{FeatureNorm} and \textit{NormRatio} for OOD detection, our method cannot achieve the best results on Resnet18 and ResNet50 architectures, and we argue that the structure of the residual block is the reason for it. In particular, we demonstrate that the position of the Batch Normalization (BN) layer\cite{ioffe2015batch} can cause the large OOD \textit{FeatureNorm} in Table \ref{tab:wobn}. Note that the order of the layer in block of ResNet\cite{he2016deep} is \textit{Conv-BN-ReLU}, but \textit{BN-ReLU-Conv} in WRN\cite{BMVC2016_87} for faster training and better accuracy. There have been reports that the BN layer cause the overconfidence\cite{guo2017calibration} and overactivation\cite{sun2021react} since it calculate the output by standardizing the input elements with moving average and moving variance of ID, and the absolute value of output elements becomes large. Then, the element-wise addition between input and BN output of the residual block (i.e., $x + f(x)$, where $f(\cdot)$ is a block operation for given input feature map $x$) will make \textit{FeatureNorm} of the output feature map larger, especially for high-complexity images that produce higher norm that is not rectified by ReLU. However, with the block that has order of \textit{BN-ReLU-Conv} produce lower OOD \textit{FeatureNorm} since the filters of the convolutional layer will be lowly activated for OOD inputs; as a result, \textit{FeatureNorm} of the output feature map is smaller for the given OOD. Thus, we argue that the block order is an important aspect for the OOD detection. In Table \ref{tab:wobn}, we show that the ResNet18 with block order of \textit{BN-ReLU-Conv} outperforms the ResNet18 with block order of \textit{Conv-BN-ReLU}.

\begin{table}[]
\centering
\resizebox{0.37\textwidth}{!}{%
\begin{tabular}{@{}ccccc@{}}
\toprule
 & \multicolumn{2}{c}{\textbf{Conv-BN-ReLU}} & \multicolumn{2}{c}{\textbf{BN-ReLU-Conv}} \\
\multirow{-2}{*}{OOD} & FPR95↓ & AUROC↑ & FPR95↓ & AUROC↑ \\ \midrule
SVHN & 7.13 & 98.65 & 6.92 & 98.68 \\
Textures & 31.18 & 92.31 & 36.89 & 91.91 \\
LSUN(c) & 0.07 & 99.96 & 0.50 & 99.82 \\
LSUN(r) & 27.08 & 95.25 & 17.93 & 96.97 \\
iSUN & 26.02 & 95.38 & 14.62 & 97.39 \\
Places365 & 62.54 & 84.62 & 49.30 & 90.76 \\
\rowcolor[HTML]{EFEFEF} 
\textbf{Average} & 25.67 & 94.36 & 21.03 & 95.92 \\ \bottomrule
\end{tabular}}
\caption{OOD detection performance comparison between two ResNet18 architectures with different block orders. \textit{Conv-BN-ReLU} refers to the basic block order of ResNet18 and \textit{BN-ReLU-Conv} refers to the block order of WRN28, which we argue that better block order for OOD detection. It demonstrates that the \textit{BN-ReLU-Conv} block order outperforms the original block order in detecting high complexity OOD images (LSUN(r), iSUN, Places365). Thus, ResNet18 and ResNet50 cannot fully leverage the proposed framework. The results are averaged over five runs.}
\label{tab:wobn}
\end{table}

%-------------------------------------------------------------------------
\section{Related Work}
\label{sec:works}

\subsection{Out-of-distribution detection}
To detect OOD samples during the inference stage with pre-trained network using cross-entropy loss, Hendrycks \etal\cite{hendrycks17baseline} proposed maximum softmax probability (MSP), based on their observation that the classifier tends to have lower confidence for the OOD sample than for the ID sample. Similarly, the ODIN\cite{liang2018enhancing}, which is an enhanced version of the MSP, applies two strategies, namely input preprocessing and temperature scaling, to separate the confidence of ID samples and the one of OOD samples, which leads to improved OOD detection performance. In the other hand, Lee \etal\cite{lee2018simple} proposed to use Mahalanobis distance between prototype feature vector, build with training data, and feature vector for the given input sample to detect OOD sample. Also, Liu \etal\cite{liu2020energy} proposed to use energy function for OOD detection. Recently, Sun \etal\cite{sun2021react} proposed a simple technique that clips activation of the produced feature based on the observation that OOD features have few outsized activations. Also, Sun\etal\cite{sun2022dice} proposed a weight selection method to select important weight of the overparameterized network, and it can separate energy score between ID and OOD. The above methods can be utilized with any off-the-shelf network. Despite the convenience that it can use by any trained network, its improvement of OOD detection performance is limited. Our method also belongs to the category of post-hoc methods, and we compare our method with other post-hoc methods in this paper.

The other branch of OOD detection aims to train the network for improving its OOD detection performance. For example, Hendrycks \textit{et al.}\cite{hendrycks2018deep} proposed the outlier exposure, which trains the network to have low confidence in outlier data examples, and found that it results in the network calibrated better and enhance OOD detection performance. Also, Papadopoulos \textit{et al.}\cite{papadopoulos2021outlier} improved outlier exposure with confidence control to improve the calibration of the network. Mac\^{e}do \textit{et al.}\cite{isomax} proposed a novel loss to replace the cross-entropy loss to follow the maximum entropy principle\cite{thomas2006elements}. The above works with outlier exposure can improve the OOD detection performance by a large margin, but its drawback is that it requires an outlier dataset, which can be hard to require in practice. There have been works that utilize self-supervised learning to improve performance. For instance, Tack \etal\cite{tack2020csi} proposed a novelty detection method based on representation knowledge of the model, which is learned by contrastive learning of visual representations. Also, Sehwang \etal\cite{sehwag2021ssd} proposed to use the Mahalanobis distance based detection using the model trained with self-supervised learning method. Recently, However, the main drawback of these methods is that it cannot use with a pre-trained network, and it may reduce the classification accuracy. In this paper, we did not compare performance with these methods since our method is a post-hoc method.

\subsection{Calibration of neural networks}
We prefer a well-calibrated network for deploying the neural network in the real world since a calibrated network will produce low confidence for the given unseen input. However, there has been a report the deep neural network, especially modern architecture, is poorly calibrated\cite{guo2017calibration} due to the overfitting and Batch Normalization. The calibration methods can be categorized into post-hoc methods, which utilize the recalibration function for pre-trained model, and pre-hoc methods, which utilize extra training procedure. The very basic way to improve model calibration using recalibration method is temperature scaling of logit\cite{guo2017calibration}. Also, Gupta \etal\cite{gupta2021calibration} proposed to utilize a recalibration function based on splines, which maps the logit to calibrated class probabilities. These methods are performed using a held-out calibration set to calculate hyperparameter of the calibration method. Calibration methods with training procedure use data augmentation \cite{thulasidasan2019mixup, yun2019cutmix} or modify training loss\cite{kumar2018trainable, mukhoti2020calibrating, karandikar2021soft}. These calibration methods may be helpful for OOD detection researches since the well-calibrated network should produce low probability for OOD.

% Expected Calibration Error (ECE; \cite{naeini2015obtaining}) or Overconfidence Error (OE; \cite{thulasidasan2019mixup}) are common metrics for measuring calibration of networks. 
\subsection{Feature of neural networks} 
It is known that the norm of the feature vector will be lower for unseen images since the feature extractor is forced to produce higher norm in the training stage\cite{vaze2022openset}. Also, there have been reports that the norm of the feature vectors represented the quality of the input image in the face recognition\cite{ranjan2017l2, qin2018convolutional, kim2022adaface}, and the norm of the feature vectors for unknown samples is often lower than those of known\cite{dhamija2018reducing, vaze2022openset}. However, it is also known that the network can be easily fooled by unrecognizable images\cite{nguyen2015deep, moosavi2016deepfool, hein2019relu} or adversarial attack\cite{madry2017towards, dong2018boosting} which means that the filter of the network can be activated by unseen image.  Also, the feature vector, which is directly used for calculating logit by dot product, is forced to be large during its training stage by cross-entropy loss\cite{xu2020feature}. Since our work utilizes the norm of the feature map, we believe that our framework can be improved by considering aforementioned works.

%------------------------------------------------------------------------
\section{Conclusion}
\label{sec:conclusion}
We propose a simple OOD detection framework that consists of two operations: (1) \textit{FeatureNorm}, which is a norm of feature map from the block, and (2) \textit{NormRatio}, which is a ratio of \textit{FeatureNorm} for given ID and OOD images. We demonstrate that the suitable block for OOD detection can be selected without access to OOD by choosing the block that produce the largest \textit{NormRatio} with ID and pseudo OOD that is generated from ID, and using \textit{FeatureNorm} of suitable block for OOD detection outperforms existing approaches. We provide empirical and theoretical analysis to help understand our framework. Extensive experiments show our framework can improve other existing methods as well. We hope our research can help to solve the overconfidence issue of neural networks.

\section*{Acknowledgement}
This work was partially supported by the ICT R\&D program of MSIT/IITP[2020-0-00857, Development of Cloud Robot Intelligence Augmentation, Sharing and Framework Technology to Integrate and Enhance the Intelligence of Multiple Robots] and by Institute of Information \& communications Technology Planning \& Evaluation (IITP) grant funded by the Korea government (MSIT) (No. 2022-0-00951, Development of Uncertainty-Aware Agents Learning by Asking Questions).

%%%%%%%%% REFERENCES
{\small
\bibliographystyle{ieee_fullname}
\bibliography{egbib}
}

\end{document}